# Left Ventricle Segmentation and Volume Estimation on Cardiac MRI using Deep Learning

Ehab Abdelmaguid, Jolene Huang, Sanjay Kenchareddy, Disha Singla, Laura Wilke, Mai H. Nguyen[1], Ilkay Altintas

University of California, San Diego
La Jolla, CA  USA

## Abstract

In the United States, heart disease is the leading cause of death for both men and women, accounting for 610,000 deaths each year [1]. Physicians use Magnetic Resonance Imaging (MRI) scans to take images of the heart in order to non-invasively estimate its structural and functional parameters for cardiovascular diagnosis and disease management. The end-systolic volume (ESV) and end-diastolic volume (EDV) of the left ventricle (LV), and the ejection fraction (EF) are indicators of heart disease.  These measures can be derived from the segmented contours of the LV; thus, consistent and accurate segmentation of the LV from MRI images are critical to the accuracy of the ESV, EDV, and EF, and to non-invasive cardiac disease detection.

In this work, various image preprocessing techniques, model configurations using the U-Net deep learning architecture, postprocessing methods, and approaches for volume estimation are investigated.  An end-to-end analytics pipeline with multiple stages is provided for automated LV segmentation and volume estimation.  First, image data are reformatted and processed from DICOM and NIfTI formats to raw images in array format. Secondly, raw images are processed with multiple image preprocessing methods and cropped to include only the Region of Interest (ROI). Thirdly, preprocessed images are segmented using U-Net models. Lastly, post processing of segmented images to remove extra contours along with intelligent slice and frame selection are applied, followed by calculation of the ESV, EDV, and EF.  This analytics pipeline is implemented and runs on a distributed computing environment with a GPU cluster at the San Diego Supercomputer Center at UCSD.

---

[1] Corresponding author:  Mai H. Nguyen, mhnguyen@ucsd.edu

# Table Of Contents

# 1. Introduction

In the diagnosis of the heart disease, one of the parameters that cardiologist examine is the volume ejected by the left ventricle. The difference of the end-diastolic volume (EDV) and end-systolic volume (ESV), which is a measure of the amount of blood that is pumped in one cardiac cycle is a parameter that is used in the diagnosis of the process. From the volume, the ejection fraction (EF) can be derived which is the ratio of (EDV - ESV)/EDV.

MRI images enable the ability to estimate the ESV, EDV, and EF. Currently, it takes imaging technicians and doctors several minutes to read the images to come to a diagnosis and it is not an easily repeatable process. Automation of parts of the process to determine cardiac parameters and/ or function can lead to faster consistent diagnosis and create a repeatable process for diagnosis.

Over the past several years, LV segmentation algorithms have been evolved but they have had limited success due to the lack of available labeled data. Over the years, more datasets have been made available publicly which has resulted in improvement in LV segmentation algorithms and more methods have been developed to help with the problem.

This report consists of several of the methods that have been used by those who have done previous work.

# 2. Related Work

Segmentation of Left Ventricle continues to be a challenging task despite significant evolution of techniques and network architectures in last ten years. [24] describes fully automated iterative thresholding method for LV segmentation. [25] talks about a deep convolutional encoder-decoder model seg-net for image segmentation and [22] uses Fully Convolutional Neural Network (FCN) for cardiac image segmentation task. U-Net architecture[8] another encoder-decoder architecture that has performed well on biomedical image segmentation tasks. In recent years several papers have been published on direct LV volume predictions without segmentation. [20] describes the approach using deep convolutional network and compares the performances of volume prediction using VGG,Google-net and Resnet architectures.Several solutions [13][14] that were presented during Data Science Bowl 2 Challenge used different algorithms and network architectures to perform LV segmentation and volume prediction.

# 3. Data Sources

Over the past few years, there has been an increasing number of publicly available cardiac MRI Images with labels that identify the contours of the Left Ventricle and other cardiac features. In 2009, the Sunnybrook Cardiac Data (SCD) was made available through the Cardiac MR Left Ventricle Segmentation Challenge. Kaggle.com organized the Second Annual Data Science Bowl (DSB) in 2015 which provided the largest set of MRI images. In 2017, the Medical Image Computing & Computer Assisted Intervention (MIACCI) organized the Automated Cardiac Diagnosis Challenge (ACDC) that made an additional set of MRI heart images publicly available. All three datasets were utilized for this effort.

There are three different types of views that an MRI Image can take in order to examine the heart. These views are the 4-Chamber view, 2-Chamber view, and the Short Axis view. For this analysis, the focus was on using the Short Axis (SAX) views. For the SAX view, a singular heart is broken up into multiple slices, each slice is the part of the heart at a different physical location. Each slice consists of multiple frames that show the heart across one cardiac cycle, temporal aspect. Figure 1, shows the depiction of the SAX View images.

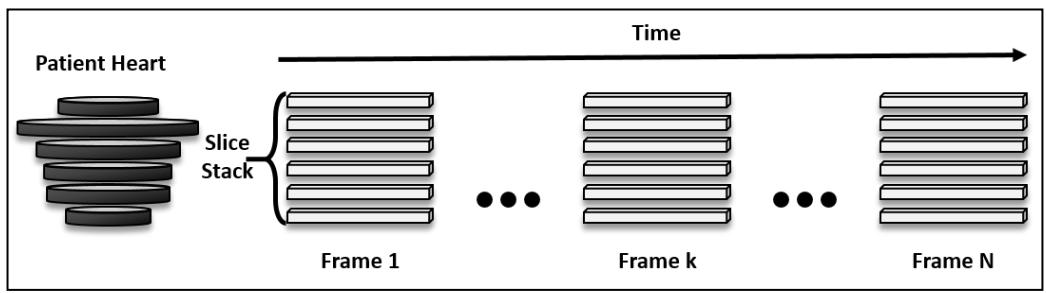

Figure 1: Patient Short Axis view; multiple slices (spatial element) consisting of multiple frames (temporal element). The number of slices per patient can vary.

The SCD data consists of 45-cine MRI images (1.6 GB) from a mixed group of patients and pathologies: healthy, hypertrophy, heart failure with infarction, and heart failure without infarction [2]. The MRI images are in the DICOM (Digital Imaging and Communications in Medicine) image format that consists of several metadata parameters about the patient and the image. For each patient record, there is a set of hand drawn contours for EDV and ESV slices. The contours were drawn by Perry Radau from the Sunnybrook Health Science Centre. The contours were available in text files that consisted of the contour points, which needed to be converted into an image. Figure 2, shows an MRI Image SAX image and the corresponding LV contour that was provided in the dataset.

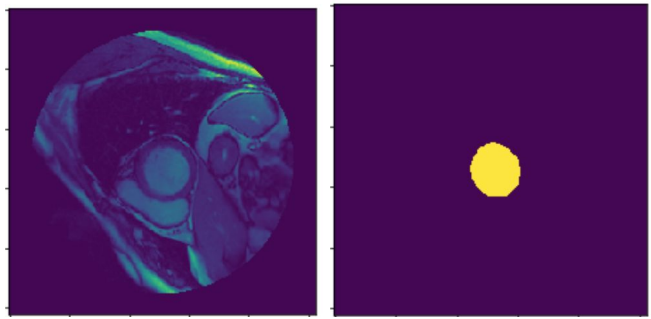

Figure 2: SCD Image and Corresponding Label

The ACDC dataset consists of SAX MRI Images for 100 patients (3.3 GB) in the NIfTI (Neuroimaging Informatics Technology Initiative) image format. Similar to the DICOM images, the NIfTI images consist of metadata about the patient and the image [3]. Each patient directory consists of a 4-D NIfTI Format Images. Contour files have been provided for the End-Systolic and End-Diastolic images for each patient. These contours were drawn to follow the limit defined by the aortic valve. This method for defining how the contours were drawn may differ from how the SCD contours were drawn as they were drawn by a different individual. Figure 3 shows a SAX Image that was provided and the corresponding label. The label shows the outer (green) and inner (yellow) LV segmentation, inner LV contour label was used in the project.

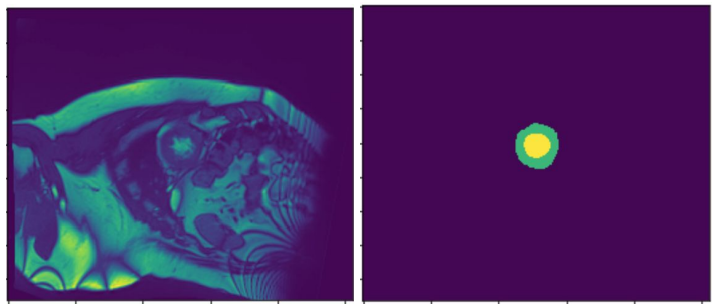

Figure 3: ACDC Image and Corresponding Label

The third dataset is the DSB dataset which consists of MRI images for 1,140 patients (100GB). The dataset includes other views of the heart, namely the 4-Chamber view and 2-Chamber view, in addition to the SAX view. For our work, however, only the SAX views were used. This dataset did not come with contour labels but rather provides an End-Systolic Volume (ESV) and End-Diastolic Volume (EDV) for each patient. These volumes are derived from the contours that are drawn from the MRI images. Dr. Anai's group from the National Heart Lung and Blood Institute drew the contours for the DSB data, which were the basis of the ESV and EDV calculations. The group's methodology for identifying the LV contours differs from the groups that drew the contours for the SCD and ACDC datasets. Dr. Anai's group looks to where one can see left ventricular muscle in order to draw the contour, resulting in a "half-moon" contour or a partial slice instead of a circular contour [4]. Figure 4, shows two SAX images and the corresponding contours that the team would draw. The first is similar to the methodology of the SCD and ACDC groups. The second produces the partial slice which the other datasets do not have.

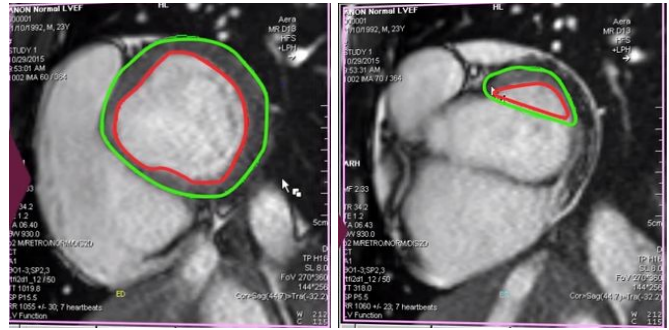

Figure 4: The red and green circles show how Dr. Anai's team would draw the contours. The right-hand image shows the contour for a frame where the aortic valve is visible. The left-hand image shows the partial slice contour that the SCD and ACDC datasets would not have provided a contour for this image or designated it as an ESF or EDF.

The SCD and ACDC datasets were used for training the U-Net model to automatically segment the Left Ventricle from the MRI image. The DSB dataset was used as the test data. There are two stages to the overall process. The first is the segmentation of the left ventricle, which was evaluated on the results of the U-Net model metrics, which included the dice similarity measure, Jaccard similarity measure, precision, recall, and others. The second step was to segment the Left Ventricle from the DSB dataset (using the trained network) and then deriving the ESV and EDV for each patient. The volume results were evaluated based on the root mean squared error (RMSE) metric.

## 4. Data Preparation

The NIfTI images needed to be converted to a similar format as the DICOM Images. The NIfTI images were broken out by the slice and the frames within the slice into 2-D numpy arrays to be used for preprocessing. The DICOM images were converted into numpy arrays for preprocessing as well. Based on the Patient Orientation metadata field, the patient orientation relative to the image plane in ACDC images is RAH (Right Anterior Head), while it is LPH (Left Posterior Head) in the DICOM Images. In order to convert the ACDC Images to the LPH orientation, the images were rotated by 180 degrees. Both the SCD and ACDC labels provided labels for more features in the heart than the LV. The labels were simplified to only provide the contour for the inner left ventricle.

Each MRI image was taken with a different MRI machine, by a different radiologist, and on a different patient. Due to these variables, the MRI images across each dataset are not consistent from one to the next. Across the datasets, the image's pixel spacing, image size, and image orientation are different. This information is stored in the metadata of the DICOM and NIfTI images. The DICOM images had more metadata parameters than the NIfTI images. The charts in Figure 5 below show the differences in the pixel spacing, image size and orientation for the three datasets. The ACDC images are the most uniform within one dataset in the sense that all ACDC images are oriented in the same way.  However, the metadata indicates that the orientation for Sunnybrook and DSB data is the same, but the orientation for ACDC images is different.

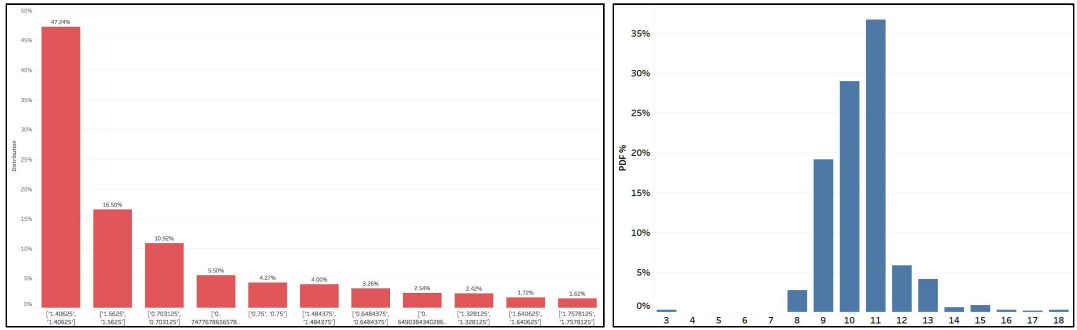

Figure 5: The graph on the left shows the different pixel spacings for all of the images within the dataset. On the right is a graph of the number of slices per patient.

Due to the differences between the images, the images need to be normalized in order to effectively and automatically segment the LV. In order to normalize these images, multiple preprocessing steps and methods were used. There are 3 Methods: Baseline, Method 1, and Method 2. Method 1 and Method 2 were derived from the third and first place winners of the Data Science Bowl competition. The Method 1 and Method 2 each have two variations which slightly alter the original image processing methods.

## 4.1. Image Preprocessing

 The Baseline Method involved rescaling the images so each pixel was 1mm x 1mm and then cropping the image from the center to either 256 x 256 or 176 x 176.

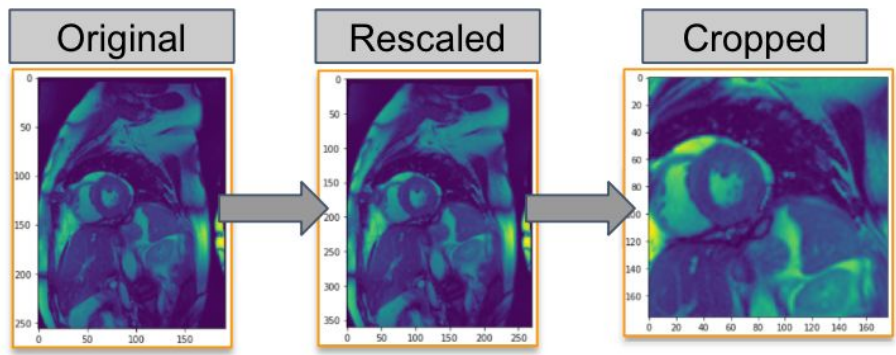

Figure 6: Baseline Normalize Method

Method 1 was derived from the third Data Science Bowl competition winner [5]. This method consisted of the following steps:

1.  **Orientation:** Orientation shift based on the DICOM InPlanePhaseEncoding metadata, which indicates the axis of phase encoding with respect to the image. A majority of the images were 'Row' oriented hence if the image was 'Col' oriented, it was flipped to be 'Row' oriented.
2.  **Rescale:** The image is rescaled based on the image's Pixel Spacing values.
    a.  Rescale with the first Pixel Spacing value in both the x and y directions
    b.  Rescale to 1mm x 1mm
3.  **Crop:** The image is cropped from the center to 256 x 256. In this project, 256x256 and 176x176 were used.
4.  **CLAHE:** CLAHE is applied to the image for each one by one tile of the image.

Method 1 Type 0 applies steps 1, 2a, 3, and 4 as described above to each image. Method 1 Type 1 applies steps 1, 2b, 3, and 4 on each image. Lastly, Method 1 Type 2 applies steps 2b, 3, and 4 on each image. The orientation switch was eliminated based on MRI Image Processing Domain expert advice. The MRI radiologist attempt to optimize the orientation of the image to capture the best picture of the heart. The orientation change was not applied to the ACDC images due to the evaluation that they consistently had the same orientation.

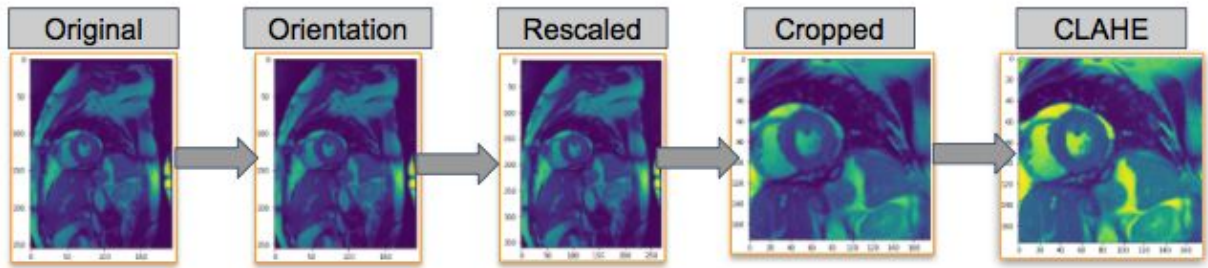

Figure 7: Method 1 Type 2b Image Normalize Steps

Method 2 was derived from the first Data Science Bowl competition winner [6], which consists of the following steps:

1.  **Orientation:** Using the ImagePositionPatient and ImageOrientationPatient metadata, rotate each image so it is oriented along a common vector. The ImagePositionPatient is the x, y, and z coordinates of the upper left corner of the image. ImageOrientationPatient is the direction cosines of the first row and the first column with respect to the patient.
2.  **Rescale**: Rescale the image to 1mm x 1mm pixel spacing.
3.  **Crop:** Crop the image from the center outward to 256 x 256 or 176 x 176.

Method 2 Type 0 applies steps 1 and 3 as described above. Method 2 Type 1 applies steps 1-3 and Method 2 Type 2 includes steps 2 and 3, eliminating the orientation step per the advice of the MRI Domain Expert. Similarly, as in Method 1, the orientation step was not applied to the ACDC images.

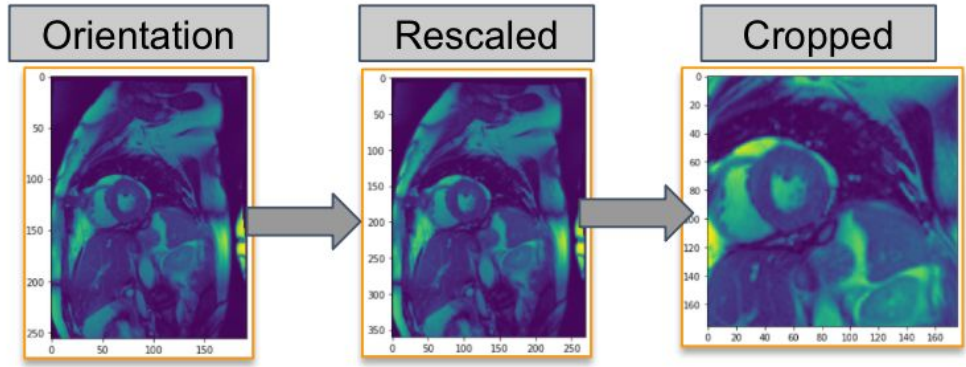

Figure 8: Method 2 Type 2 Image Processing

One of the goals was to assess which preprocessing steps contributed to the success of the segmentation of the LV. In order to evaluate this, several experiments were performed on training the U-Net on the different types of normalized images. The results will be discussed in the Findings section.

Note that all preprocessing methods, except for Baseline, also include pixel intensity normalization, which is described in the section on Model Tuning below.

## 4.2. Region of Interest Identification

In order to identify the left ventricle Region of Interest (ROI), two approaches were used. The LV ROI crops down the MRI image to focus the LV, which is expected to improve the results of the segmentation task in terms of processing and accuracy. By using an ROI image as the input into the U-Net, there would be less noise in the image. The LV has four distinct features within the MRI Image that were exploited in order to identify the ROI:
1. The heart cavity containing the LV is near the center of the MRI Image
2. The frequency at which the LV moves is unique compared to the frequencies of other heart muscles
3. There is a degree of pixel variance around the LV muscle
4. The LV is circular in shape.

The first approach included using the rescaled images and cropping to 176 x 176 with the assumption that the LV would be within that area, as shown in Figure 9. This was utilizing the first feature of the LV within the MRI Image, stated above.

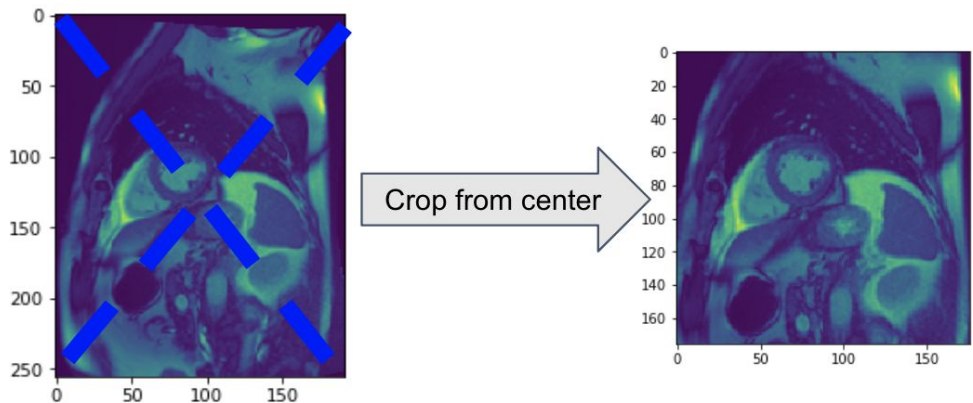

Figure 9: The rescaled image cropped from the center to be 176 x176

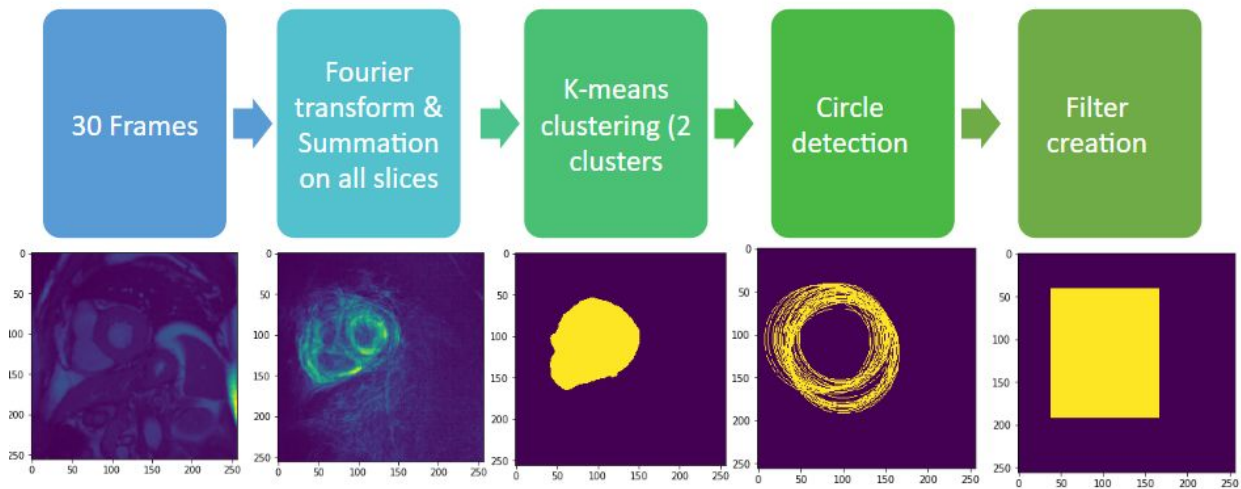

Figure 10a: The steps of the second approach to ROI.

The second approach included using the images, rescaled to 1mm by 1mm pixel spacing, and applying additional steps in order to further reduce the ROI space, as described in Figure 10a.

This approach used Fourier time transform to detect the first harmonic due to the motion by creating a 3D array from 2D image frames.  A 3D array is (2D image, T), where T is the time component. The first harmonic for each slice of MRI image was calculated.  Then all of the first harmonics were added together across all slices to increase the magnitude of the region of LV as it is the region with most motion across all slices.

Next, k-means clustering, with k=2, was performed on the image resulting from summing the first harmonic of all slices.   High values which indicated motion, where the left ventricle was present, were in cluster w, while low values where no motion for left ventricle was in another cluster.

The next step was circle detection using the Hough transform. The Hough transform is a feature extraction technique original used to find lines in an image, but has been extended to detect positions of arbitrary shapes such as circles. The Hough transform is applied to the image resulting from k-means clustering to identify circles. The range of radii in pixels (15 to 64), along with the final number of circles to retain at the end (30), are specified as parameters.

Then, min/max X & Y positions of edges of circles detected were extracted to create a rectangular filter of pixel values of one (yellow) surrounded with zero (black) as the binary image. The rectangular filter was expanded by 10% on all sides, based on the number of pixels, to ensure the LV region is included. The binary image was applied to the original image to pass only the region of interest, as shown in Figure 10b.

Although the second approach was more accurate to detect ROI, it did not perform well for volume calculation, as shown in the Results section below, possibly due to the black border surrounding ROI.

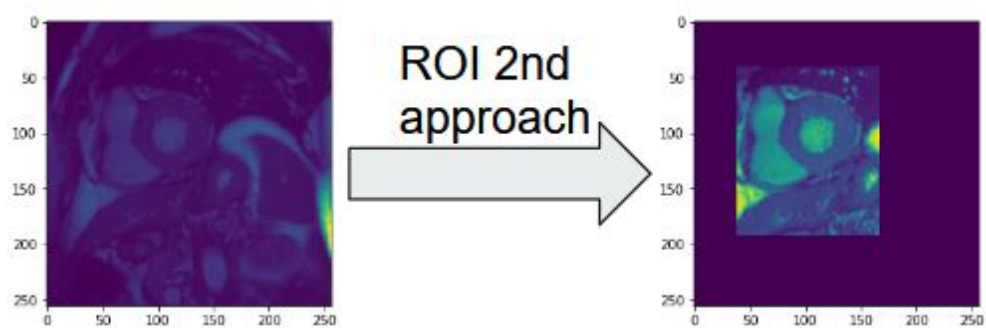

Figure 10b: Final result of ROI

# 5. Left Ventricle Segmentation

The task of LV segmentation is to generate an image mask that separates LV (area within endocardial wall) from rest of the structures in a SAX image. The segmentation task has two parts. 1) Localization: Identify the LV in SAX image, 2) Pixelwise classification: identify the pixels belonging to LV and label them as 1s and rest of the pixels as 0s (background class)

## 5.1. Why U-Net?

Numerous papers and blogs[15, 16] show that U-Net architecture achieves better performance and accuracy than regular CNN in the segmentation of biomedical images for following reasons:

1. The regular CNNs focus mainly on classification tasks, take an image as input and output a single class label. However, as explained in the previous section the biomedical image segmentation requires one to classify each pixel in the image and U-Net works well for pixel-wise prediction tasks. This localized prediction is very important for biomedical image segmentation tasks.

2. U-Net has been proven to work very effectively even with fewer training samples (few hundred)
3. U-Net does not have a fully connected layer at the end. This relaxes the restrictions on input image size

## 5.2. U-Net Architecture

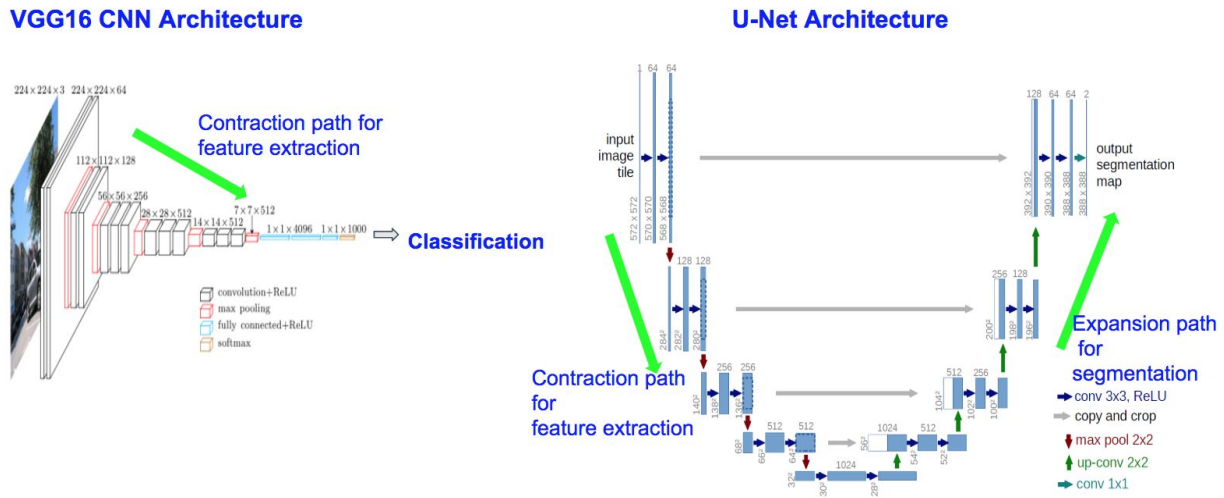

Figure 11: Traditional CNN[27][28] Vs U-Net[8] Architecture

### 5.2.1. Traditional convolutional network architecture

The VGG16 architecture[27][28] in the diagram above depicts the traditional convolutional neural net architecture. Feature extraction is done hierarchically through a series of convolution filters followed by non-linear activation functions such as Relu, sigmoid. A Series of max-pooling layers are used to downsample the output and create the contraction path. Extracted features are then fed to a fully connected neural network for the final classification task.

### 5.2.2. U-Net architecture

U-Net architecture is shown in the Figure 11. U-Net is an encoder-decoder type of network[8]. It consists of a contracting path on the left that follows the typical architecture of a convolution neural network with a set of convolution filters followed by ReLU activation function and max pooling layer to downsample the output after each stage. The contraction path does the same job as a traditional CNN and it is responsible for feature extraction. The major architecture change in the U-Net is the expansive path on the right side which consists of upsampling of the channels followed by a series of convolutional filters. This path is responsible for localization and segmentation. One important step in upsampling is the concatenation of feature map from contracting path at each stage as shown by gray arrows in the figure. This will allow the propagation of contextual information to higher resolution layers. The final layer is 1x1 convolution layer that maps each 64 component feature vector to the desired number of classes.

## 5.3. Performance Evaluation criteria

The biomedical image segmentation performance can be measured using similarity functions. To compute the similarity between ground-truth contours and predictions dice similarity score and Jaccard similarity score were used.

The Dice Similarity Coefficient is computed using the formula :

$$DSC = 2\,(Y\,true \cap Y\,pred)/(|\;Y\,true| + |\;Y\,pred|\;)$$

$$Dice\;Loss = 1 - DSC$$

The Jaccard Similarity coefficient is the same as IOU (Intersection over Union), and is computed using the formulas:

$$JSC = (Y\,true \cap Y\,pred)/(Y\,true \cup Y\,pred) = (Y\,true \cap Y\,pred)/(|\;Y\,true| + |\;Y\,pred| - (Y\,true \cap Y\,pred))$$

$$Jaccard\;Loss = 1 - JSC$$

In addition to the above two similarity measures, other metrics used were precision, recall, and f1-scores to measure the performance of the segmentation.

## 5.4. Model Tuning

As shown in the diagram below, the parameters to tune U-Net model for better performance falls into three categories. 1. Model Parameters, 2. Image Parameters, 3. Model Hyper parameters. The results of performance tuning are discussed in the following sections.

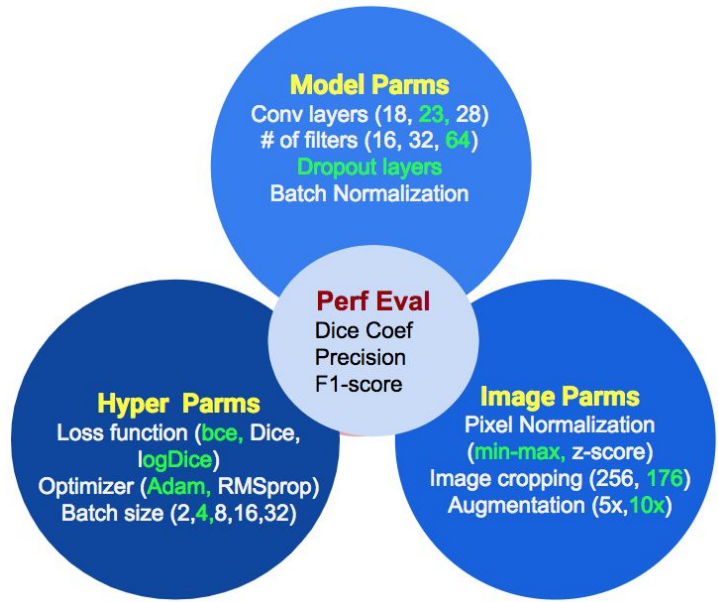

Figure 12: Model performance tuning

### 5.4.1. Number of convolution layers.

Model performance was evaluated with U-Net with different sizes. The results in Table 1 shows that  U-Net with 23 convolution layers is more optimal with training time and accuracy. Adding more layers did not improve the accuracy of the model.

| Conv Layers | Training Weights | Training Time (100 Epochs, 2GPUs) | Performance (Dice Coef) |
|---|---|---|---|
| 18 | 1.9M | 4 Hr | 92-93% |
| 23 | 31M | 6Hr | 94-96% |
| 28 | 32M | 7.5Hr | 94-96% |

Table 1:  Model Performance with Convolutional layers

### 5.4.2. Number of Filters

Model performance was evaluated with 16, 32 and 64 filters at the first convolution layer on the decoder path. The number of filters is doubled at every subsequent convolution layer (64->128->256->512->1024). The model with 64 filters at the first convolution layer showed best performance.

### 5.4.3. Batch Normalization

We experimented with adding a Batch Normalization layer between every Convolutional layer and Activation layer in both encoding and decoding paths.  The default parameters for the Batch Normalization layer, as implemented in Keras, were used.  We did not see any significant changes in the Dice coefficient, accuracy, precision, recall, or F1-score.  No significant changes in training time were observed, either.

### 5.4.4. Dropout layers

We also experimented with adding Dropout layers before each Upsampling layer in the encoding path.  Dropout levels of 20% and 50% were tested.  Although our tests indicated that adding dropout layer did not improve the performance of the model, we did include dropout layers in our final model since many of the solutions we studied suggested performance improvement with them.  In our final model, a dropout layer with 50% was added after each convolution layer in the encoding path, for a total of four dropout layers.

Table 2 captures the results of experiments conducted on models with/without dropout layers for two different runs.

| Attributes | No Dropout1 | No Dropout2 | Dropout1 | Dropout2 |
|---|---|---|---|---|
| dice_coef | 0.956 | 0.956 | 0.953 | 0.952 |
| precision | 0.965 | 0.955 | 0.957 | 0.952 |
| recall | 0.947 | 0.957 | 0.950 | 0.952 |
| f1_score | 0.956 | 0.956 | 0.953 | 0.952 |

Table 2: Model Performance evaluation with dropout layers

Overfitting was not encountered even when the dropout layers were not included.  Figures 13 and 14  show the learning history in both cases.

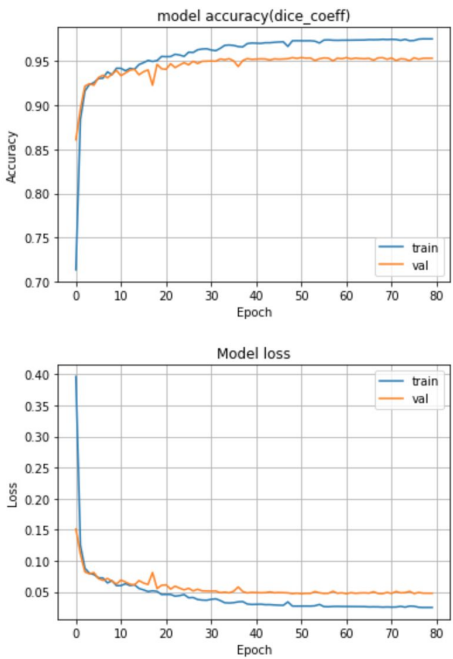

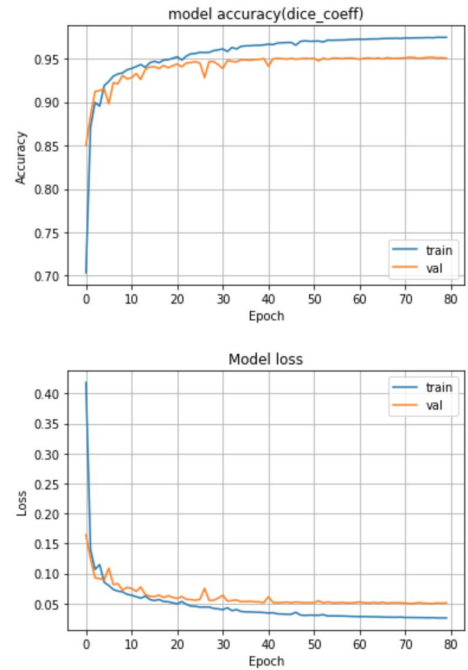

Figure 13: Learning History with no dropout layers        Figure 14: Learning History with Dropout layers

### 5.4.5. Hyperparameter Tuning

● Loss function.
    Model performance was evaluated with following loss functions
    1.    Binary Cross Entropy
    2.    Dice Loss
    3.    Log (Dice loss)
    4.    Binary Cross Entropy +  Dice loss

The model performance did not vary significantly with these loss functions. The precision/F1-score were within 94-96% range. log(dice loss) gave a slightly better performance. The results are shown in Table 3.

| Attributes | Dice Loss | BCE+Dice | Log Dice | BinaryCrossEntropy |
|---|---|---|---|---|
| dice_coef | 0.951 | 0.948 | 0.956 | 0.954 |
| precision | 0.952 | 0.945 | 0.965 | 0.959 |
| recall | 0.950 | 0.951 | 0.947 | 0.952 |
| f1_score | 0.951 | 0.948 | 0.956 | 0.955 |

Table 3: Model performance evaluation with loss functions

- Optimizers: Model training performance was evaluated with following two optimizers provided by Keras
  - Adam optimizer
  - RMS prop

  Adam optimizer showed better convergence
- Learning rate:  The experiments showed that the learning rates needed to be between 1E-4 and 1E-5. It also depended on the batch size.
- Batch size:  Model performance was evaluated with the following batch sizes:  4, 8, 16, 32, and 64.  Batch size of 4 yielded best results.

## 5.4.6. Pixel Intensity Normalization

The pixel depth of MRI images in our dataset is 16 bits. However, the images have very low dynamic range with pixel intensities between 0 and 4000. Pixel Intensity normalization improved the segmentation performance. We tested two methods

a.  Min-max normalization : Each pixel value $X_i$ is normalized using formula  $(x_i - x_{min})/(x_{max} - x_{min})$

b.  Z-score normalization : Each pixel value $X_i$ is normalized using formula $(x_i - x_{mean})/x_{stdvid}$

Both normalization methods showed similar performance gains. Our final experiments used min-max normalization.

## 5.4.7. Image Cropping

We tested with two different sizes for image cropping:  256x256 and 176x176. The cropping was done from the center and after applying pixel spacing normalization to the original image. With SAX images we were able preserve the ROI containing the LV region even after images were cropped to 176x176 pixels. The images with 2Ch and 4Ch views may lose part of the ROI with such aggressive cropping.

We chose 256x256 because many of the SAX images had at least one dimension of 256 pixels. We chose 176x176 because other approaches we studied used a smaller image size such as 180x180. Another reason to choose 176 is because it is divisible by 16.  Our final U-net  model has 4 downsampling (maxpooling) layers in the encoder path. Selecting the image size that is divisible by 16 will ensure that there is no cropping of image due to rounding in each of these downsampling steps. This also eliminates the cropping or padding  steps required to match the image sizes during concatenation steps in decoder path of U-net.

The smaller image size of 176x176 yielded slightly better performance results on the validation set and much faster training time.  With 2 GPUs and 100 epochs, 256x256 images took 8 hours to complete training, while 176x176 images tooks 6 hours, a 25% reduction.

## 5.4.8. Data Augmentation

We augmented the training set using transformations rotate (0-90 degrees), shift (0-5% of image height/width) and zoom (0.95 - 1.05). Figure 15 shows results of some representative augmentation operations.

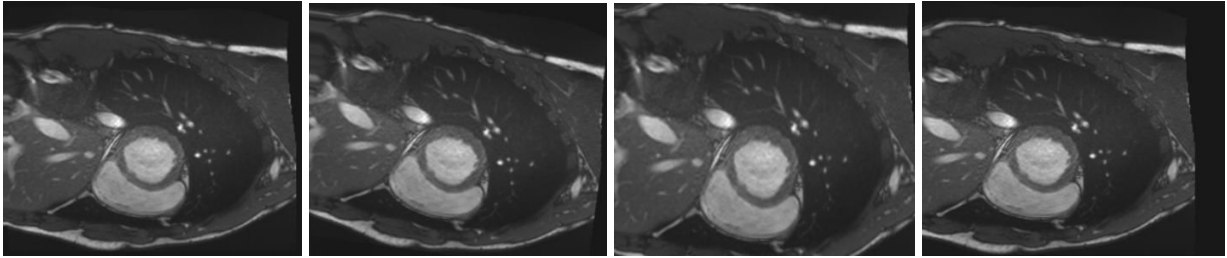

Figure 15: From left to Right: Original Image, Image Zoomed, Image Rotated, Image Shifted

Data augmentation results at different levels (0X, 4X, and 10X) are shown in Table 3C. Augmentation at 10X gave the most performance gains, and that is what was used for the rest of our experiments.  We also found that augmentation also helped to reduce run-to-run variation.

| Image augmentation | Training set size (Number of images) | Performance range (Dice Coef) |
|---|---|---|
| No Augmentation | 2000 | 91-93.5% |
| 4x Augmentation | 2000+8000 | 92-95.5% |
| 10x Augmentation | 2000+20,000 | 94.5-96.5% |

Table 3C:  Model Performance with image augmentation

## 5.4.9. Run-to-Run Variation

The experiments on different loss functions, dropout layers etc. did not show a significant difference in terms of performance. The minor differences in the performances could be just from run-to-run variations. To measure run-to-run variation all parameters were kept constant and the training was run multiple times. There was a 1-4% performance variation between the runs, as indicated in Table 4.

| Attributes | Run-1 | Run-2 | Run-3 |
|---|---|---|---|
| dice_coef | 0.956 | 0.956 | 0.954 |
| precision | 0.965 | 0.955 | 0.957 |
| recall | 0.947 | 0.957 | 0.950 |
| f1_score | 0.956 | 0.956 | 0.954 |

Table 4: Run-to-run variation

## 5.5. Final U-Net Model

As explained in previous sections, increasing the convolutional layers from 18 to 23 improved the performance from 92% to 95%. Increasing the depth further to 28 layers did not show any improvement. Hence it was decided to use 23 layer U-Net model for the final training task. Table 5 shows the configuration of different layers in the implementation.

| Layer (type) | Output Shape | Param # | Connected to |
|---|---|---|---|
| input_1 (InputLayer) | (None, 176, 176, 1) | 0 | |
| conv2d_1 (Conv2D) | (None, 176, 176, 64) | 640 | input_1[0][0] |
| conv2d_2 (Conv2D) | (None, 176, 176, 64) | 36928 | conv2d_1[0][0] |
| max_pooling2d_1 (MaxPooling2D) | (None, 88, 88, 64) | 0 | conv2d_2[0][0] |
| conv2d_3 (Conv2D) | (None, 88, 88, 128) | 73856 | max_pooling2d_1[0][0] |
| conv2d_4 (Conv2D) | (None, 88, 88, 128) | 147584 | conv2d_3[0][0] |
| max_pooling2d_2 (MaxPooling2D) | (None, 44, 44, 128) | 0 | conv2d_4[0][0] |
| conv2d_5 (Conv2D) | (None, 44, 44, 256) | 295168 | max_pooling2d_2[0][0] |
| conv2d_6 (Conv2D) | (None, 44, 44, 256) | 590080 | conv2d_5[0][0] |
| max_pooling2d_3 (MaxPooling2D) | (None, 22, 22, 256) | 0 | conv2d_6[0][0] |
| conv2d_7 (Conv2D) | (None, 22, 22, 512) | 1180160 | max_pooling2d_3[0][0] |
| conv2d_8 (Conv2D) | (None, 22, 22, 512) | 2359808 | conv2d_7[0][0] |
| max_pooling2d_4 (MaxPooling2D) | (None, 11, 11, 512) | 0 | conv2d_8[0][0] |
| conv2d_9 (Conv2D) | (None, 11, 11, 1024) | 4719616 | max_pooling2d_4[0][0] |
| conv2d_10 (Conv2D) | (None, 11, 11, 1024) | 9438208 | conv2d_9[0][0] |
| dropout_1 (Dropout) | (None, 11, 11, 1024) | 0 | conv2d_10[0][0] |
| up_sampling2d_1 (UpSampling2D) | (None, 22, 22, 1024) | 0 | dropout_1[0][0] |
| conv2d_11 (Conv2D) | (None, 22, 22, 512) | 2097664 | up_sampling2d_1[0][0] |
| concatenate_1 (Concatenate) | (None, 22, 22, 1024) | 0 | conv2d_8[0][0]<br>conv2d_11[0][0] |
| conv2d_12 (Conv2D) | (None, 22, 22, 512) | 4719104 | concatenate_1[0][0] |
| conv2d_13 (Conv2D) | (None, 22, 22, 512) | 2359808 | conv2d_12[0][0] |
| dropout_2 (Dropout) | (None, 22, 22, 512) | 0 | conv2d_13[0][0] |
| up_sampling2d_2 (UpSampling2D) | (None, 44, 44, 512) | 0 | dropout_2[0][0] |
| conv2d_14 (Conv2D) | (None, 44, 44, 256) | 524544 | up_sampling2d_2[0][0] |
| concatenate_2 (Concatenate) | (None, 44, 44, 512) | 0 | conv2d_6[0][0]<br>conv2d_14[0][0] |
| conv2d_15 (Conv2D) | (None, 44, 44, 256) | 1179904 | concatenate_2[0][0] |
| conv2d_16 (Conv2D) | (None, 44, 44, 256) | 590080 | conv2d_15[0][0] |
| dropout_3 (Dropout) | (None, 44, 44, 256) | 0 | conv2d_16[0][0] |
| up_sampling2d_3 (UpSampling2D) | (None, 88, 88, 256) | 0 | dropout_3[0][0] |
| conv2d_17 (Conv2D) | (None, 88, 88, 128) | 131200 | up_sampling2d_3[0][0] |
| concatenate_3 (Concatenate) | (None, 88, 88, 256) | 0 | conv2d_4[0][0]<br>conv2d_17[0][0] |
| conv2d_18 (Conv2D) | (None, 88, 88, 128) | 295040 | concatenate_3[0][0] |
| conv2d_19 (Conv2D) | (None, 88, 88, 128) | 147584 | conv2d_18[0][0] |
| dropout_4 (Dropout) | (None, 88, 88, 128) | 0 | conv2d_19[0][0] |
| up_sampling2d_4 (UpSampling2D) | (None, 176, 176, 128 | 0 | dropout_4[0][0] |
| conv2d_20 (Conv2D) | (None, 176, 176, 64) | 32832 | up_sampling2d_4[0][0] |
| concatenate_4 (Concatenate) | (None, 176, 176, 128 | 0 | conv2d_2[0][0]<br>conv2d_20[0][0] |
| conv2d_21 (Conv2D) | (None, 176, 176, 64) | 73792 | concatenate_4[0][0] |
| conv2d_22 (Conv2D) | (None, 176, 176, 64) | 36928 | conv2d_21[0][0] |
| conv2d_23 (Conv2D) | (None, 176, 176, 1) | 65 | conv2d_22[0][0] |

=================================================================================================
**Total params: 31,030,593**
**Trainable params: 31,030,593**

Table 5: 23 Layer U-Net model summary

The model has 23 convolution layers, 10 layers on the contraction path (left side of U-Net) and 13 layers on the expansion path on the right side of the U-Net. 3x3 kernels with depths from 64 to 1024 were used as kernels in each convolution layer.  Input feature maps were padded to prevent loss of any rows or columns after convolution. Dropout layers were added to the encoder path of the U-Net. The model consisted of a total of 31,030,593 trainable parameters (filter weights). BinaryCrossEntropy  and dice-coefficient were used as loss functions. Layer 23 is 1x1 convolution with a sigmoid activation function and it outputs the pixel values between zero and one for the segmented mask.

## 5.6. Training and Validation Approach

The data set included 2800 SAX images from 145 patients (800 images from 45 patients from Sunnybrook + 2000 images from 100 patients from ACDC). The images were split in to training and test sets using 90:10 ratio.  The training set was further split into Train: Validation set with ratio 80:20. Splitting was done based on patients so that all images belonging to a patient are in a single set (train, validation, or test).   Input images were cropped from the center to fit sizes 256x256 and 176x176.

Note that the 2800 images were those at end-diastole (ED) and end-systole (ES) since only ED and ES frames come with contours.  Note also that for both datasets, some ES images do not come with a corresponding contour labels, especially for the slices at the apex and base.  Such patient records in Sunnybrook dataset have only contour label files for ED (and not for ES) for the slices at the apex and base.  For ACDC, for such patient records, the contour label files for images at apex or base have no voxels representing LV contour. In other words, a corresponding blank contour were provided if the image does not show any LV presence.  We included these images with no LV contours from ACDC dataset into our training data set. We generated blank contour files as labels for such images.

The Keras built-in image augmentation functionality was used to augment images with the following affine transformations:  shift, rotate, and flip.

# 6. Segmentation - Post Processing

## 6.1. Removal of Extra Contours

The segmentation model occasionally predicted more than one contour. This typically occurs in cases when the Right Ventricle was prominent in the image or when there is another circular cardiac feature present. Two methods were developed to remove the contour/contours that did not represent the LV.

The first method is based on the size of the contours. When multiple contours are predicted for one image, the contour with the largest area (i.e., the contour mask containing the most pixels) is selected as the LV contour. The remaining contours are removed by setting their pixel values to zero to indicate background. This method sometimes selected incorrect contour for the LV, as shown in the Figure 16 below. The second image shows the output of segmentation which produced two contours. The third image shows the output of this method, which selected the incorrect contour.

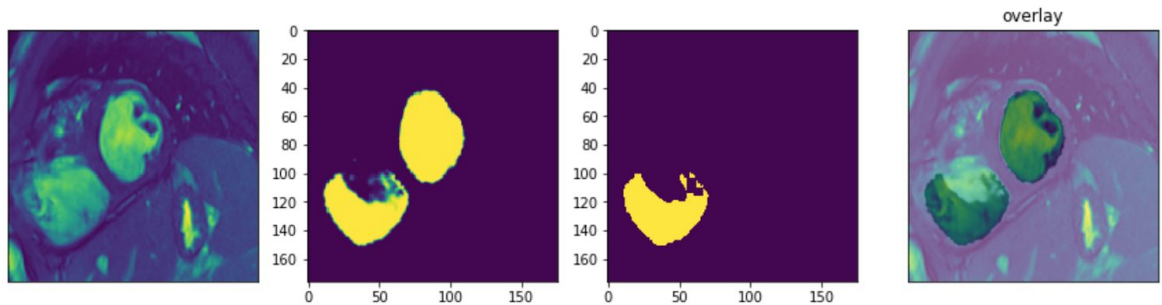

Figure 16: Results of removal of extra contours using the first method

The second method uses the entire patient record (i.e., all frames and slices for each patient) to identify the LV contour. The underlying assumption for this method is that the locations of the center of LV in all images, across all slices of a patient record, lie within a small distance from each other.

The segmentation model takes the entire patient record as input in the form of a 4D array. The first dimension represents the image number, the second and third dimensions represent row and column size of each image, and the fourth dimension represents the number of color channels (which equals 1 for monochrome MRI images). The segmentation output is also a 4D array with the same dimensions. In order to identify the center of the LV across all of the slices, the pixel values of the predicted contour images are added across all images (along the first dimension in the 4-D array). This results in a 2-D array where each element represents the concentration of LV contour pixels in the image plane, as shown in the heatmap mask image in Figure 17 below. The center of the LV is then determined by identifying locations of highest pixel concentrations. For each image in a patient record, when more than one contour is predicted, the contour that encompasses the center of the LV is retained and rest of the contours are removed. If none of the predicted contours encompasses the center of the LV, then all contours are removed, resulting in a blank contour image.

Figure 17 shows results of contour removal using the second method. The first case shows a single predicted contour, so no contour removal was necessary. The next two cases shows multiple contour predictions. Using this method, the top contour was selected, which was the correct LV contour. In the last case, identified center of the LV was not within any of the predicted contours, so all of those contours were eliminated, resulting in a blank contour image to indicate that no contour was predicted for this image.

The second method selected the correct LV contour more consistently than the first method, and thus, was included in our final solution.

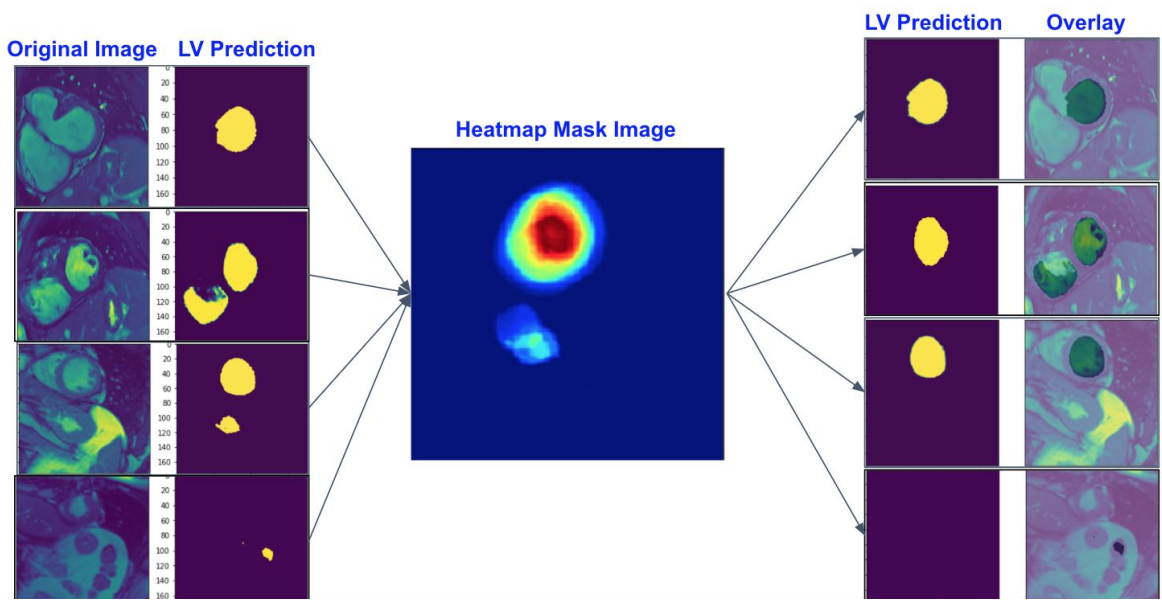

Figure 17: Results of removal of extra contours using the second method

# 7. Volume Calculation

The flow diagram of training and validation for segmentation and volume calculation is shown in the Figure 18. As shown in the figure, there are two separate sub-processes: one for segmentation, and one for volume calculation. The segmentation sub-process is shown in the top part and uses Sunnybrook and ACDC data. The best performing segmentation models are then used in the volume calculation sub-process in the bottom part. The segmentation models are applied to the DSB "training" data (which includes the training and validation datasets provided by DSB, consisting of 700 images) to segment the LV in those images and compute EDV and ESV.  The best volume models are then determined and applied to the DSB test data (consisting of 440 images) to get the final EDV and ESV results.

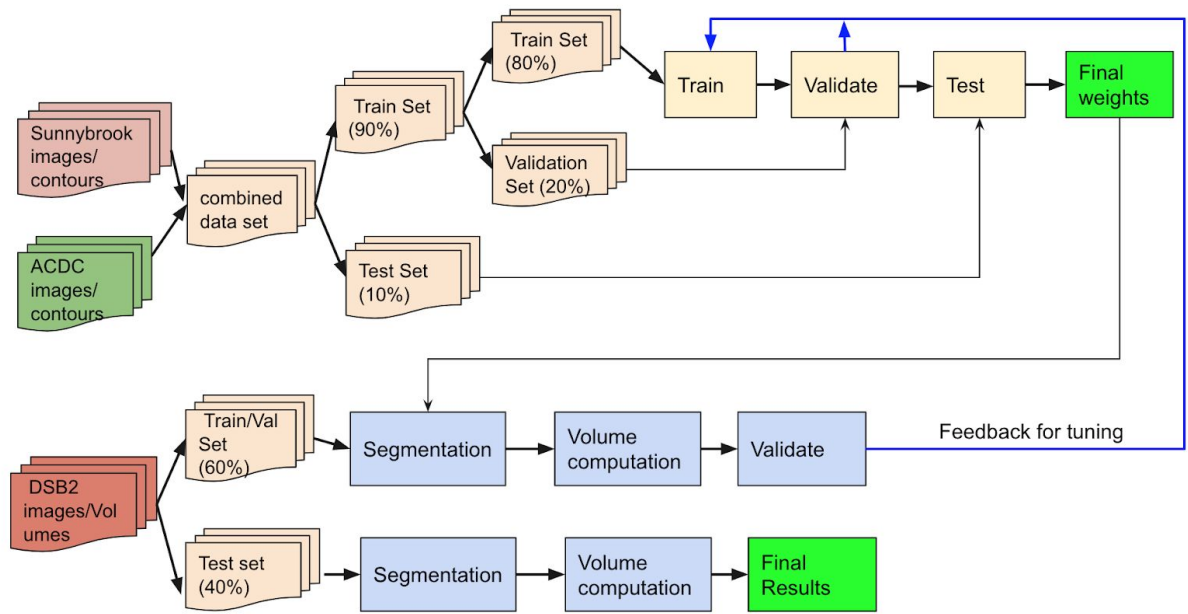

Figure 18: Training and Validation process flow for segmentation and volume calculation

For each image that was input to the U-Net, an output image was produced where each pixel value was the likelihood that the pixel was part of the LV contour. The pixel values were binarized based on a 0.5 threshold. This threshold was chosen because the likelihoods for a single prediction were at the extremes, either close to zero or close to one.  From this binarized image, the physical area can be derived since the image's pixel spacing and size are known.

$$A_i = \text{F} * (\text{p} * \text{I})^2$$

where $A_i$ is the physical area of the ith slice, F is the fraction of pixels that are part of the LV contour, p is the pixel spacing, and I is the image length (specified as number of pixels) [7].

The area of the heart is treated as a truncated cone, which is the sum of the volumes between neighboring slices of the heart.  The next step is to calculate the location, $L_i$, of each slice along the z-axis of heart. Then the slices are ordered based on their location on the z-axis. For each slice, two frames are selected to be included in the volume calculation. The frame with the minimum area is the End-Systolic Frame (ESF) for that slice and the frame with the maximum area is the End-Diastolic Frame (EDF).

Then volume between two neighbor slices is a truncated cone, whose volume is

$$V_i = (A_i + A_{i+1} + \sqrt{A_i A_{i+1}})(\tfrac{L_i - L_{i+1}}{3}).$$

where $A_i$ is the area of the ith slice.

L, the slice location on the z-axis, is determined by using the DICOM parameters Image Position Patient (IPP) and Image Orientation Patient (IOP). IOP is a 1x3 vector and IOP is a 1x6 vector where IOP[0:3] is the orientation in the row direction and IOP[4:6] is the orientation in the column direction. L is calculated as follows:

$$L = IPP \cdot (IOP[0:3] \times IOP[4:6])$$

where $\times$ is the cross product, and $\cdot$ is the dot product. This projects the locations for each slice into the same plane. This calculation for L is used instead of the slice location parameter in the DICOM metadata (Slice Location Attribute) since we found that this parameter is not reliably, and in some cases, not correctly populated.

Previous work substituted the arithmetic mean into the formula above for $V_i$ based on empirical analysis [7]. We also use this in our calculation, and this changes the volume between two slices to be

$$V_i = (A_i + A_{i+1})\left(\tfrac{L_i - L_{i+1}}{2}\right).$$

The overall volume of the heart can be computed by:

$$V = \sum_{i=0}^{N-1} V_i = (A_i + A_{i+1})\left(\frac{L_i - L_{i+1}}{2}\right)$$

where N is the number of slices within the the heart. This calculation is performed for the ESFs and EDFs of the patient's heart to compute the ESV and EDV, respectively.

Figure 19 shows the overall volume calculation process. For each image that was identified to be part of End-Systolic or End-Diastolic, the images are sorted based on their slice location. Then using the predicted area of the LV, the ESV and EDV are calculated.

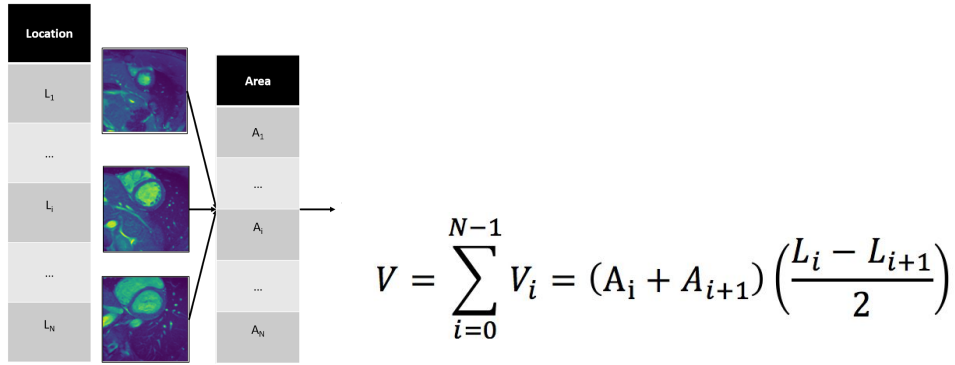

$$V = \sum_{i=0}^{N-1} V_i = (A_i + A_{i+1})\left(\frac{L_i - L_{i+1}}{2}\right)$$

Figure 19: The process for deriving the patient's volume

There are several patient records that have multiple sets of frames for one slice in the heart. Typically, this means that the radiologist went back to that heart location to retake the MRI Images of that slice. Due to this, if a patient record had multiple sets of frames for one slice, the set of frames that was taken last by the radiologist was used for the volume calculation.

Occasionally, there were cases where all the frames for a single slice were predicted to be zero. If this occurred between $S_{i+1}$ and $S_N$ where $S_i$ is the $i^{th}$ slice within the patient record, that slice was eliminated and the volume was extrapolated from the slice above and below the slice that was removed.

Based on the information from the domain expert in this project, the first slice in ES or ED should have a smaller area than the second slice. Similarly, at the end of the heart, the last slice should have a smaller area than the slice before it. If the area of the first or last slice were larger than the second or the second to last slice, respectively, those slices were removed from the volume calculation. This assumes that those slices are not valid LV contours.

# 8. Results

## 8.1. Optimal Image Preprocessing Technique

In order to examine which image preprocessing technique performed the best on the segmentation task, the model performance results were evaluated for each preprocessing technique across four different models. Note that Method 1 Type 0 and Method 2 Type 0 are not included in the table since those preprocessing techniques yielded worse results than the other techniques and were subsequently dropped. These were the initial U-Net models in order to determine which preprocessing method would be continued to use. Based on Precision, Recall, and the F1 Score, the differences between the normalization methods is negligible. This can been seen in the Table 6, 7, and 8.

| | Precision | | | | |
|---|---|---|---|---|---|
| **Image Preprocessing** | Baseline | Method 1 Type 1 | Method 1 Type 2 | Method 2 Type 1 | Method 2 Type 2 |
| **DICE** | 0.931 | 0.912 | 0.941 | 0.959 | 0.950 |
| **Augmentation** | 0.952 | 0.950 | 0.950 | 0.957 | 0.962 |
| **Drop Layers** | 0.924 | 0.897 | 0.900 | 0.963 | 0.965 |
| **Augmentation & Drop Layers** | 0.940 | 0.954 | 0.959 | 0.964 | 0.962 |

Table 6: Precision metric

| | F1 Score | | | | |
|---|---|---|---|---|---|
| **Image Preprocessing** | Baseline | Method 1 Type 1 | Method 1 Type 2 | Method 2 Type 1 | Method 2 Type 2 |
| **DICE** | 0.941 | 0.920 | 0.945 | 0.944 | 0.953 |
| **Augmentation** | 0.951 | 0.948 | 0.954 | 0.943 | 0.952 |
| **Drop Layers** | 0.939 | 0.920 | 0.934 | 0.944 | 0.954 |
| **Augmentation & Drop Layers** | 0.949 | 0.949 | 0.952 | 0.944 | 0.952 |

Table 7: F1 score metric

| | Recall | | | | |
|---|---|---|---|---|---|
| **Image Preprocessing** | Baseline | Method 1 Type 1 | Method 1 Type 2 | Method 2 Type 1 | Method 2 Type 2 |
| **DICE** | 0.952 | 0.940 | 0.949 | 0.930 | 0.955 |
| **Augmentation** | 0.950 | 0.939 | 0.940 | 0.930 | 0.942 |
| **Drop Layers** | 0.954 | 0.947 | 0.963 | 0.926 | 0.942 |
| **Augmentation & Drop Layers** | 0.950 | 0.943 | 0.945 | 0.925 | 0.943 |

Table 8: Recall metric

Due to these small differences, the team examined the volume calculations for each of the normalization methods with each model. The volume was used to determine the evaluation of which preprocessing method would be used moving forward with the model experiments in order to optimize the volume calculation.

| Volume | Baseline | | | Method 1 Type 1 | | | Method 1 Type 2 | | | Method 2 Type 1 | | | Method 2 Type 2 | | |
|---|---|---|---|---|---|---|---|---|---|---|---|---|---|---|---|
| | ESV (ml) | EDV (ml) | EF | ESV (ml) | EDV (ml) | EF | ESV (ml) | EDV (ml) | EF | ESV (ml) | EDV (ml) | EF | ESV (ml) | EDV (ml) | EF |
| Dice | 21.71 | 25.68 | 9.0% | 27.73 | 33.87 | 10.0% | 20.14 | 27.04 | 9.0% | 20.6 | 24.5 | 10.00% | 26.39 | 26.34 | 9.00% |
| Augmentation | 15.82 | 19.84 | 9.0% | 15.49 | 19.58 | 8.0% | 16.53 | 21.02 | 8.0% | 17.97 | 21.34 | 8.00% | 15.53 | 22.23 | 8.00% |
| Drop Layers | 23.75 | 25.25 | 8.0% | 15.24 | 38.9 | 8.0% | 30.95 | 41.35 | 10.0% | 29.89 | 34.76 | 9.00% | 32.08 | 32.68 | 9.00% |
| Augmentation & Drop Layers | 18.82 | 29.26 | 9.0% | 31.34 | 20.13 | 10.0% | 18.47 | 25.24 | 9.0% | 16.36 | 19.14 | 9.00% | 15.6 | 22.32 | 8.00% |

Table 9: Volume results for different preprocessing methods

The baseline results for ESV, EDV, and EF are consistently better across the four models that were evaluated. Due to these results, the team moved forward with the Baseline Image Processing image for further analysis.

After several enhancements with the U-Net model hyper-parameters, the volumes results improved with the Baseline Image Processing techniques. However, there were several image that the predictions were incorrect for due to the similarity between the pixels. Due to this, CLAHE was added to the Baseline Image Processing Pipeline, making it the same as Method 1 Type 2.

## 8.2. ROI Results

The ROI method was applied to the preprocessed Baseline images that were cropped to 176 x 176. Two of the above models were used in order to examine the effectiveness of the ROI in the image preprocessing step. The precision, F1 Score, and recall results for the two models are in shown in Table 10. These results are within the same space as without the ROI.

| | Precision | F1 Score | Recall |
|---|---|---|---|
| **Augmentation** | 0.951 | 0.948 | 0.944 |
| **Augmentation & Drop Layers** | 0.951 | 0.948 | 0.946 |

Table 10: ROI segmentation results

When the predicted results of both models were used in the volume calculation, the volume results were worse with ROI processes added in the image preprocessing steps, as seen in

Table 11. Due to these results, the team chose to eliminate the ROI steps from the methodology when computing the volume.

|  | ESV | EDV | EF |
|---|---|---|---|
| **Augmentation** | 64.43 ml | 75.18 ml | 36% |
| **Augmentation & Drop Layers** | 59.45 ml | 81ml | 34% |

Table 11: ROI volume results

## 8.3. Segmentation Results

The top ten performing segmentation models with various model parameters are shown in the Table 12.  Note that for all models, pixel intensity normalization and pixel spacing to 1mm-by-1mm were applied.  Several tests were run for each model, and performance numbers from the best run are shown in the table.

| Id | Parameters | dice_coef | jaccard_coef | precision | recall | f1_score |
|---|---|---|---|---|---|---|
| 1 | 10x Augmentation, bce loss, CLAHE normalization | 0.957 | 0.917 | 0.961 | 0.955 | 0.958 |
| 2 | 10x Augmentation, bce loss, CLAHE norm, batch normalization | 0.957 | 0.918 | 0.962 | 0.956 | 0.959 |
| 3 | 4x Augmentation, log_dice loss | 0.956 | 0.915 | 0.965 | 0.947 | 0.956 |
| 4 | 10x Augmentation, log_dice loss, dropout layers | 0.955 | 0.914 | 0.957 | 0.953 | 0.955 |
| 5 | 10x Augmentation, log_dice loss, CLAHE norm, dropout layers, batch norm | 0.955 | 0.913 | 0.96 | 0.949 | 0.955 |
| 6 | 4x Augmentation, bce loss | 0.954 | 0.911 | 0.959 | 0.952 | 0.955 |
| 7 | 10x Augmentation, log_dice , CLAHE norm, dropout layers | 0.954 | 0.912 | 0.958 | 0.95 | 0.954 |
| 8 | 4x Augmentation, dice loss | 0.953 | 0.909 | 0.953 | 0.952 | 0.953 |

| 9 | 10x Augmentation, bce loss, CLAHE norm, dropout layers | 0.953 | 0.91 | 0.952 | 0.96 | 0.956 |
|---|---|---|---|---|---|---|
| 10 | 4x Augmentation, bce+dice loss | 0.951 | 0.907 | 0.956 | 0.948 | 0.952 |

Table 12: Model performance evaluation on segmentation task

## 8.4. Volume Results from Top Six Segmentation Models

Table 13 below captures the volume results obtained using the top six segmentation models from Table 12. There was not a single model that showed best performance segmentation as well as ESV/EDV and EF computation. Some models performed well with ESV and some with EDV. Model 1 had best segmentation performance, however the volume results were worse on this model compared to other models. Model 6 from Table 12 was ranked 6th in segmentation performance but gave best performance for ED volume. Similarly Model 2 was ranked 2nd in segmentation performance and showed best performance on EV volume and EF.

Results in Table 13 were for the DSB 'train' data, which included data from the DSB training and validation datasets. Although this data was not used to train any models for the segmentation task, we refer to this as the 'train' data since we use it to determine the best volume models to use for the final volume estimation on the DSB test data.

| Id | Parameters | Segmentation | Volume Results | | |
|---|---|---|---|---|---|
|  |  | Dice_coef | ESV RMSE | EDV RMSE | EF RMSE |
| 1 | 10x Augmentation, bce loss, CLAHE normalization | 0.957 | 13.75 ml | 15.22 ml | 9% |
| 2 | 10x Augmentation, bce loss, CLAHE norm, batch normalization | 0.957 | 14.41ml | 14.47ml | 8% |
| 3 | 4x Augmentation, log_dice loss | 0.956 | 16.61ml | 19.4 ml | 9% |
| 4 | 10x Augmentation, log_dice loss, dropout layers | 0.955 | 14.56 ml | 15.68 ml | 8% |
| 5 | 10x Augmentation, log_dice loss, CLAHE norm, dropout layers, batch norm | 0.955 | 15.77 ml | 18.23 ml | 10% |
| 6 | 10x Augmentation, bce loss, CLAHE norm, dropout layers | 0.953 | 12.40 ml | 17.01 ml | 8% |

Table 13: Volume estimation results on DSB 'train' data using top 6 segmentation models

## 8.5. Final Volume Calculation Task

As described in the previous section, some models performed well on EDV and some on ESV. The final volume calculation model was built with two separate paths for EDV and ESV computation. Each path can use either one model or an ensemble of top performing models.

We experimented with ensembles by combining the best three EDV models from Table 13 to create an EDV ensemble, and similarly with an ESV ensemble. In each ensemble, individual models were combined in two ways: (1) each model's contour predictions were rounded, then a majority voting was applied to combine all models' predictions into the final predictions to determine which pixels were within the contour mask; (2) all models' predictions were averaged together, then rounded to provide the final contour predictions. The first approach gave better results. However, both ensemble approaches gave worse results in terms of EF RMSE. So in our final setup, we used single models to estimate EDV and ESV.

The process flow is shown in Figure 20.

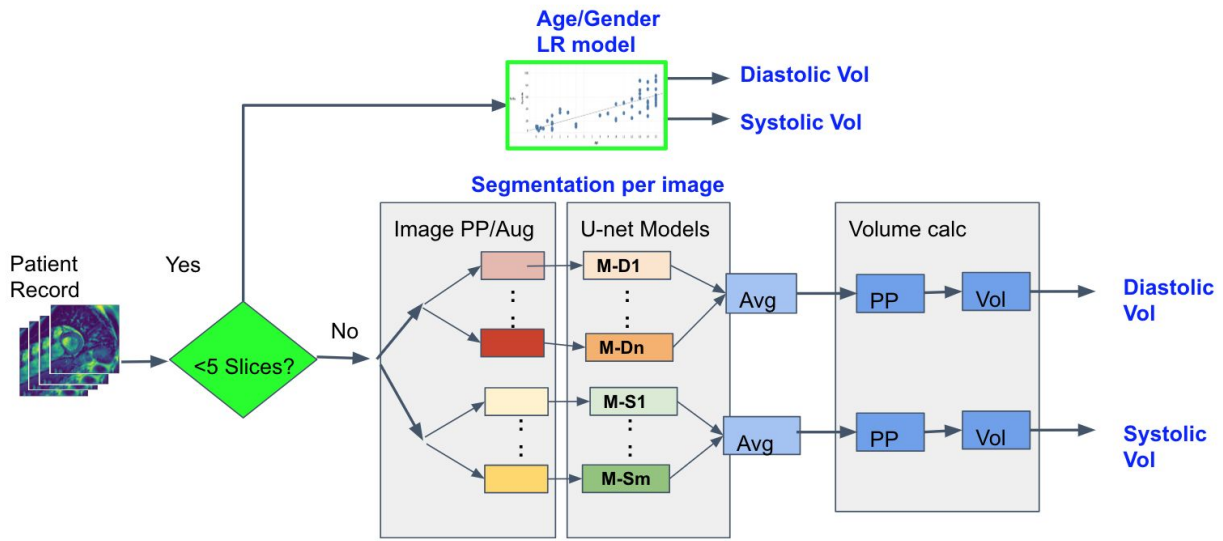

Figure 20: Final volume calculation model

## 8.6. Models for Edge Cases

The error with volume calculation was high with specific patient records that had fewer than 5 slices, and hence a linear regression model was used for patient age/gender for ESV/EDV computation for these specific cases, similar to an approach for the Data Science Bowl[26]. There were two patients with this scenario.

The linear model is based on analysis of DSB training data set (volume labeled) with gender and age. Model $R^2$ is about 0.6 with $P < 0.001$ In the table below, x represents the patient's age[26].

| Age | Gender | Systole | Diastole |
|---|---|---|---|
| **Under 16 years** | **Male** | 4.69x | 10.8x + 9 |
| | **Female** | 2.41x + 15 | 7.61x + 22 |
| **16 years and above** | **Male** | 75 ml | 181 ml |
| | **Female** | 53.6 ml | 144 ml |

Table 14: Gender Age Model

Another scenario where a linear regression model was used was when the volume prediction was low and below pre-specified thresholds (2.3 ml for ESV and 5 ml for EDV). These thresholds were determined from the training data: The smallest ESV value from the training data was 4.6; if the predicted ESV value for a patient was half of this (i.e., 2.3), then the linear model was used to determine ESV, bypassing the U-Net and volume calculation, as depicted in Figure 20. Similarly for EDV. Three patients in the test dataset met the ESV criterion, and thus, the linear model was used to determine ESV for these patients. No patients met the EDV criterion; that is, there was no test case for which EDV was predicted as less than 5 ml.

## 8.7. Volume Results

In order to compute the ESV and EDV for the DSB test data, two different prediction models were used to determine the LV contours. In training the best ESV results were derived from LV contour predictions from the U-Net model that used 10 fold augmentation, drop layers, and the binary cross entropy loss function (Model 6 from Table 13). The model that produced the best EDV results was the U-Net model with 10 fold augmentation, batch normalization, and binary cross entropy loss function (Model 2 from Table 13).

The results for the ESV RMSE, EDV RMSE, and EF RMSE on the DSB test data are in the Table 15. These results did not improve upon the work that was done previously by the top 4 winners of the Data Science Bowl. Possible reason could be due to the fact that the DSB dataset's LV contours were drawn differently than the contours for the training dataset used, SCD and ACDC. The third place DSB winner used the SCD data and hand labeled data to match the way Dr. Anai's group drew the contour labels [5]. The fourth place DSB winner used the SCD dataset and two additional datasets from previous challenges in their training [7].

| ESV RMSE | EDV RMSE | EF RMSE |
|---|---|---|
| 15.1 ml | 16.5 ml | 9.4% |

Table 15: Volume estimation results on DSB test data

Figure 21 below shows the Actual vs Predicted results for the ESV, EDV, and EF. The histograms show the span of the L-1-norm for each patient prediction. The model is over predicting more the ESV and is evenly over predicting and under predicting for EDV. For the EF, the results are over predicting the EF.

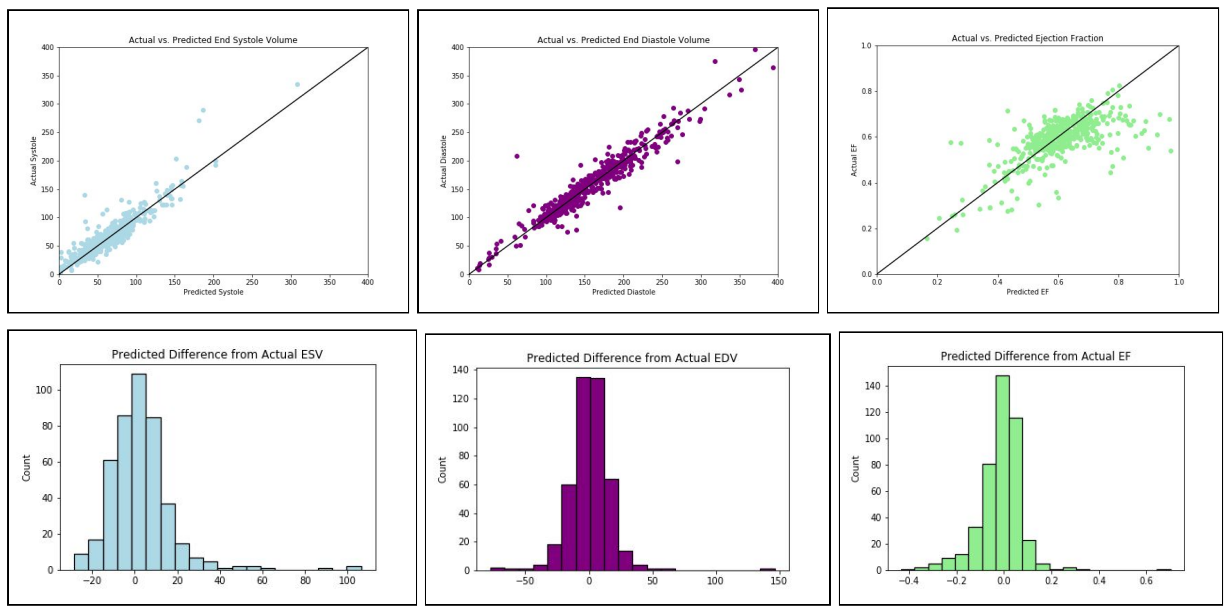

Figure 21: The top row is the Actual vs Predictions for ESV, EDV, and EF. The second row is the distribution of the difference between the actual and predicted values for ESV, EDV, and EF.

Figure 22 shows a confusion matrix (between actual vs predicted results) for the different classes of heart failure based on the EF. The heart failure classes were determined based on the EF value[29]. Based on the results, 70% of the cases were predicted correctly.

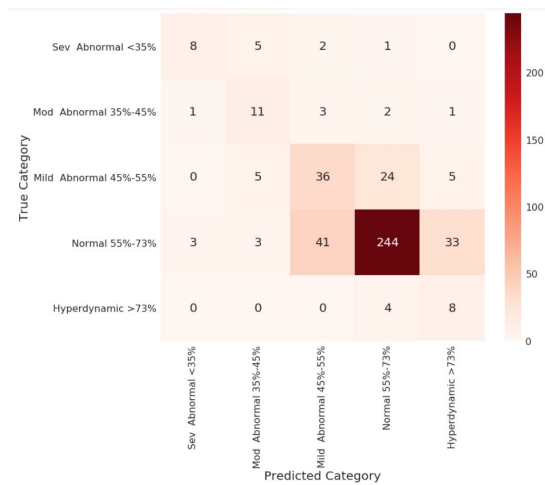

Figure 22: Confusion matrix of the the predicted category and the true category.

# 9. Findings and Lessons Learned

## 9.1. Findings

- Pixel intensity normalization and contrast normalization improve the segmentation results.
- Augmenting training set with images with no LV improved segmentation results on slices at the apex and base of the heart.
- Due to limited data, the U-Net model is sensitive to the image's orientation. Augmenting the training data with affine transforms reduces the sensitivity.
- The loss functions (dice coefficient, log loss dice, and binary cross entropy) used during training were chosen due to the fact that the dataset was highly class imbalanced.
- In order to optimize the ESV and EDV, two different models were used for the predictions.
- A linear regression model was used in order to estimate the volumes for patients who has a limited number of slices (less than five)  in their record and if the volume predictions were less than 2.3 ml in ESV and 5 ml in EDV.
- Across all the models, the End-Systolic Volume estimation constantly performed better compared to the End-Diastolic Volume estimation.

## 9.2. Lessons Learned

- Segmentation Post-Processing (removal of extra contours, removal of duplicate slices, etc.) is needed in order to compute the volume based on the segmentation predictions.
- Volume results may vary based on different contour labeling methods used by the experts in training masks, especially on the slices at apex and base of heart.

- The addition of the ROI filtering performs equally as well in the segmentation task as without ROI filtering, but the volume results are not on par with the Non-ROI filtering volume results.
- Eight GPUs with 300 GB of memory and 20 TB of storage space is an ideal resource configuration
- Due to the possible instability of the pod, it is essential to have a persistent storage in the form of a rook volume or NFS mount point
- In order to save the model weights, a dynamic way to switch between multi-gpu and a serial base model was important due to a limitation in Keras.
- The flexibility of setting GPU related variables in the training script can speed up the time spent training the model.
- The versions of the Tensorflow and cuDNN libraries need to be in sync.

## 10. Next Steps

There are several experiments that can be performed in order to improve the results. Below lists some next steps that can be taken to improve the volume results.

- Increase the set of MRI Images in the training set that would not be used for the volume calculation
- Use the second ROI approach without the zero padding or limit the amount of zero padding to a smaller area
- Estimate the volume directly without segmentation
    - This can be used in the cases where there are few slices
- Use the 2-Chamber and 4-Chamber MRI views to identify the apex and base locations of the heart
- Develop a methodology to determine which slices should be kept in the volume calculation

## 11. Conclusion

The U-Net model performed well at segmenting the left ventricle from a SAX MRI Image. The dice similarity score on best models was 95%. This performance is on par with the work that has been previously done in Data Science Bowl 2 challenge. Augmenting the train-set with affine transforms, contrast normalization and adding images with zero contours from ACDC dataset to train-set improved the performance of the model, especially in dealing with noisy images with zero LV contours at apex and base of heart.

Even with best performing segmentation models the results of volume calculation did not match the top performers of the DSB2 challenge. The ESV/EDV and EF RMSE of the model is 5 ml off from the top performing models in DSB2, possible reason could be that this is due to the differences in how the volume was calculated by each competitor. Throughout all of the experiments and the training sets, the volume estimation for ESV was consistently better than for EDV.

# 12. Solution Architecture, Performance and Evaluation

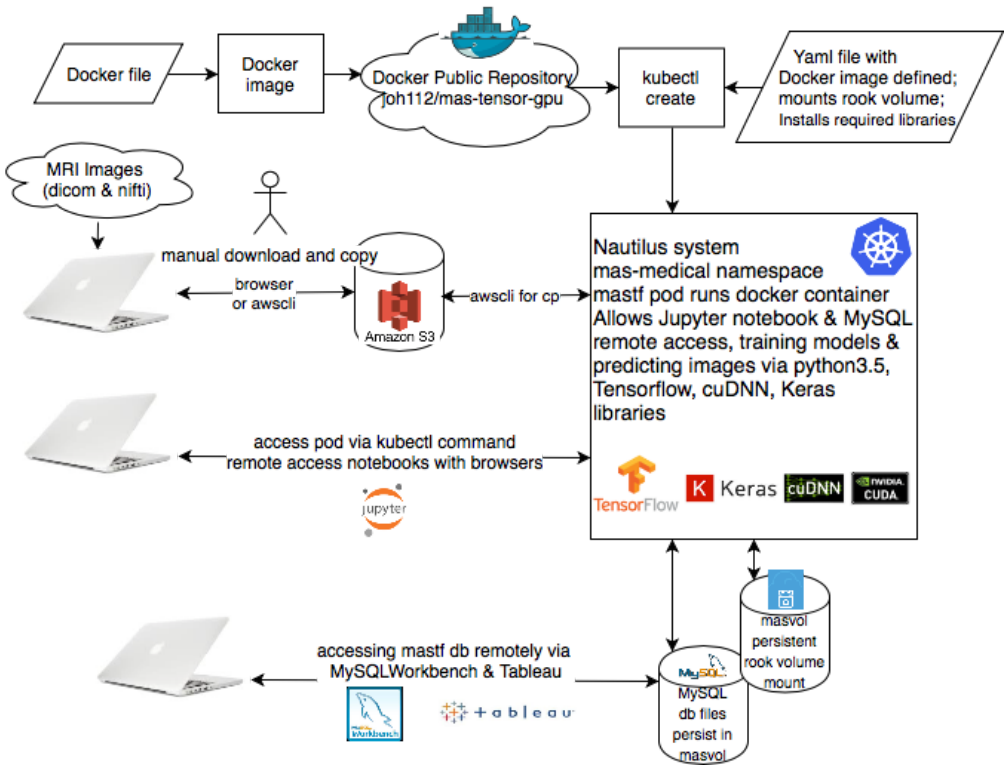

Figure 23: Infrastructure Architecture

## 12.1. Infrastructure Architecture

The infrastructure consists of several components and setups (details in Appendix C) :

**Docker**
A Docker image was built locally and pushed to the Docker repository for the defined yaml file to access when it's deployed via kubectl command. The Docker image consists of sdsc/words-tensorflow-py3 Docker image as a base.

## Kubernetes

Pod set up - A single pod, mastf was deployed to Nautilus system (SDSC) and was used concurrently by the team.   A rook volume, masvol for persistence storage and cache mount recovery purpose in the event of unforeseen pod failures were set up.

## Nvidia CUDA

Nvidia CUDA was configured to enable GPU usage in the Keras libraries.

## Amazon S3

S3 object storage was used to store the data on cloud for added level of persistence in addition to rook volume on kubernetes.

## Jupyter Notebook server set up

Most of the code was written using jupyter notebook, python 3.5.

## Github access via pod

Git clone was done from the pod to sync the code from github.  In order to keep the code up to date in github, all the codes were copied over to the laptop local area before updating to the github account.  Possible enhancement in the future is a cron setup for automatic upload.

## MySQL

MySql was installed on the pod directly to store the results in relational form for better analysis and to be used for visualization via tableau.

## Tableau

After port forwarding the MySQL 3306 port to a local port, the DB connection can be setup in Tableau to access the mastf DB directly.

A dashboard using tableau has been created to show multiple models performance at different stages, these stages are model training and volume calculation.

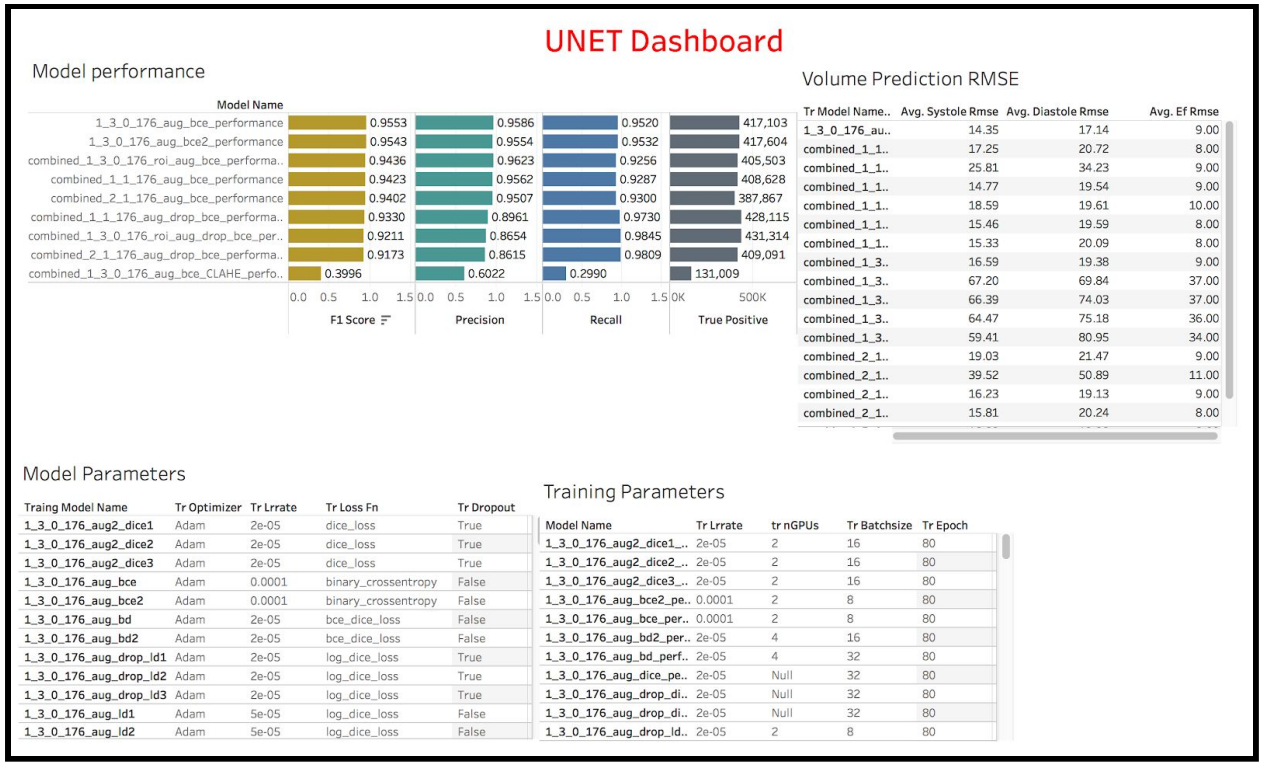

Figure 24: Results Dashboard

The dashboard consists of 4 views that covers Model performance, Model volume prediction performance, Hyperparameters and training parameters Model performance chart has a view with multiple model names with key performance indicators as F1 score, Precision, Recall & True positive value. Volume Prediction table show Systole RMSE , Diastole RMSE and Ejection fraction RMSE. Model parameters table shows Optimizer used in training, learning rate and dropout usage ( True or False).Training parameters table shows Number of GPUs used during training, Batch size and number EPOCHs. Using the dashboard helps to analyze the different models performance end to end.

## 12.2. Scalability

1.  Scalability with Execution environment
    a.  CPU and GPU Execution environment: the training, validation or testing of the model can be done on either CPU or GPU environment.
    b.  Execution parallelism: It was necessary to run several experiments in parallel to evaluate the performance of the model with different preprocessing techniques

and also to tune the hyper parameters of the model.Scalable execution environment was designed in a way that up to eight tasks could run in parallel, each with dedicated GPUs. Keras can execute using multiple GPUs. The GPU data parallelism work with the following steps. Keras divide the input batch into same number sub batches as the GPUs.Second a model copy will be applied to each GPU.Lastly, the results after training or testing are then aggregated as a single output using CPU.

2. Scalability with programming platform

   a. Scalability across multiple ML frameworks.The platform should be able to run on various proven backend numerical computation libraries, machine learning libraries such as tensorFlow, Theano, CNTK etc.

3. U-Net model scalability

   a. Scalability with volume of training data. The model meets acceptable performance even with limited training data.

   b. Scalability with training process: The model should have capability to be re-trained incrementally whenever new training data is available.

The robustness requirements are captured below :

1. Various backup options to save/restore the information

   a. Saving and restoring of execution environment

      i. Docker image built and pushed to Docker repository

      ii. Kubernetes yaml file was saved in github and gitlab

      iii. Github and gitlab

   b. Saving and restoring data

      i. Backups in AWS S3

      ii. Rook volume, masvol

2. Data storage formats optimized for specific data types

   a. The outputs of each of the different stages are stored either using image formats, or numpy arrays to minimize the processing overhead and storage memory

   b. The backup files are in zip format

# 13. Acknowledgements

The authors would like to thank the following individuals for fruitful discussions related to this project:  Marcus Bobar, Eric Carruth, Dmitry Mishin, Evan Muse, and Gary Cottrell.

This work was supported in part by NIH grants 1UL1 TR001114 and 1UL1 TR002550-01, and NSF-1730158 for Cognitive Hardware and Software Ecosystem Community Infrastructure (CHASE-CI).